\documentclass[sigconf]{acmart}
\usepackage{booktabs}
\usepackage{multirow}
\usepackage{array}

\usepackage{algorithm}
\usepackage{algpseudocode}

\usepackage{arydshln}
\usepackage{graphicx}

\usepackage{tabularx}
\usepackage{booktabs}
\usepackage{enumitem}

\AtBeginDocument{%
  }

\setcopyright{acmlicensed}
\copyrightyear{2026}
\acmYear{2026}
\acmDOI{XXXXXXX.XXXXXXX}
\acmConference[KDD '26]{The Undergraduate Consortium at KDD 2026}{August 9--13, 2026}{Republic of Korea}
\acmISBN{979-8-4007-2258-5/2026/08}

\begin{document}

\title{RAPID: Layer-Wise Redundancy-Aware Pruning and Importance-Driven Token Merging for Efficient ViT}

\author{Kyumin Choi}
\email{dinoboy22@hufs.ac.kr}
\affiliation{%
  \institution{Hankuk University of Foreign Studies}
  \city{Seoul}
  \country{South Korea}
}

\author{Ikbeom Jang} 
\authornote{Corresponding author.}
\email{ijang@hufs.ac.kr}
\affiliation{%
  \institution{Hankuk University of Foreign Studies}
  \city{Seoul}
  \country{South Korea}
}


\begin{abstract}

Vision Transformers (ViTs) achieve strong performance but suffer from high computational costs due to quadratic self-attention complexity. Although token reduction techniques such as pruning and merging mitigate this, they typically overlook how representations evolve across network depth. We propose RAPID, a depth-aware token reduction framework that adapts reduction strategies to the layer-wise characteristics of token representations. The primary methodological contribution is a bifurcated strategy: in shallow-to-middle layers, RAPID employs a redundancy-similarity aware pruning metric to eliminate over-represented local patterns. As features transition to global semantic concepts in deeper layers, the framework shifts to an importance-similarity aware merging mechanism. This stage leverages classification (CLS) token attention weights to protect semantically critical tokens while fusing less important but similar neighbors. Empirical validation on ImageNet-1K using ViT and DeiT architectures demonstrates that RAPID establishes a superior accuracy-compression Pareto frontier compared to plug-and-play baselines such as ToMe and ToFu. RAPID is particularly robust in aggressive compression regimes, achieving up to 4.29\% higher accuracy than ToMe at extreme reduction rates. Our framework provides a training-free template for optimizing vision models by aligning reduction strategies with hierarchical feature evolution. Code is available at \url{https://github.com/labhai/RAPID}.

\end{abstract}

\ccsdesc[500]{Computing methodologies~Neural networks}
\ccsdesc[500]{Computing methodologies~Computer vision}
\ccsdesc[500]{Computing methodologies~Machine learning}

\keywords{Token Merging, Token Pruning, Depth-Aware Token Reduction, Model Compression, Vision Transformer, Efficient AI}


\maketitle

\section{Introduction}
The emergence of Vision Transformers (ViTs) in computer vision has enabled higher performance than conventional CNN-based approaches, owing to their weaker inductive bias and superior ability to capture global context~\cite{dosovitskiy2020vit, vaswani2017attention}. However, because ViTs structurally require self-attention to be applied to all tokens, they suffer from a major drawback: compared with CNNs, they require substantially more computation. As a result, it has been difficult to deploy Vision Transformers in scenarios where computational resources are limited or fast inference is critical, which has motivated a wide range of model compression and acceleration efforts.

One major line of research on lightweight Vision Transformers focuses on reducing the complexity of the model architecture itself. This direction aims to alleviate the high computational cost of ViTs by redesigning the backbone with structures that require fewer operations and less memory. Representative examples include architectures such as MobileViT~\cite{mehta2022mobilevit} and EfficientViT~\cite{liu2023efficientvit}. However, a key limitation of this line of work is that these models generally require pre-training from scratch. Consequently, they may be less flexible in leveraging the rapidly growing number of newly emerging high-performance pretrained models.

Another major direction is to improve efficiency by reducing the number of tokens at the token level. This approach is particularly effective because it directly exploits the defining characteristic of ViTs: an input image is represented as a sequence of patch tokens. A representative example is DynamicViT~\cite{rao2021dynamicvit}, a token-pruning-based method that progressively removes less important tokens to reduce unnecessary computation and improve inference efficiency. However, it requires an fine-tuning process to determine which tokens should be preserved or discarded, and because tokens are completely removed, information loss may occur. On the other hand, EViT reduces the number of tokens not by simply discarding less important ones, but by fusing them into important tokens~\cite{liang2022evit}. Compared with pruning, this strategy can mitigate information loss. However, EViT still requires additional training, making it less directly applicable to arbitrary pretrained models. This line of research is further advanced by ToMe~\cite{bolya2023tome}, which enables plug-and-play acceleration through token merging of similar tokens. Because ToMe can be applied to various pretrained Vision Transformers without additional training, it demonstrates strong practical utility.

However, despite its strong practical utility, ToMe also has important limitations. First, in the early layers of a Vision Transformer, tokens tend to encode local patterns and texture-like information rather than global or semantic information~\cite{park2022how}. As a result, when two tokens are fused at this stage, their meanings may not combine properly and can instead become distorted~\cite{kim2024token}. Second, in the deeper middle-to-late layers, where tokens have already learned more global contextual information, merging without considering token importance may dilute the information carried by semantically important tokens, causing critical features to be weakened and thereby significantly affecting the final prediction~\cite{kim2024token, liang2022evit, rao2021dynamicvit}. This issue can become even more severe in later layers, where the number of tokens has already been substantially reduced and each remaining token therefore carries a larger share of the representation.

To address these limitations, we propose RAPID, a depth-aware token reduction framework that adaptively applies different reduction strategies according to the layer-wise characteristics of token representations. In shallow-to-middle layers, where tokens mainly encode local patterns and low-level signals, RAPID performs redundancy-aware pruning to remove highly overlapping tokens while minimizing representational distortion and preserving token diversity~\cite{jeddi2025similarity, kim2024token}. In middle-to-late layers, where tokens contain richer semantic context, RAPID performs importance-aware merging by preserving important tokens and merging only less important but highly similar ones, thereby reducing computation while retaining information from both tokens~\cite{liang2022evit}. By jointly considering redundancy, importance, and similarity rather than relying on similarity alone, RAPID achieves consistent improvements across various ViT backbones, especially under aggressive token reduction settings where preventing performance degradation is most critical.

\section{Related Works}

\subsection{Architectural Methods}
Research on lightweight Vision Transformers through architectural redesign focuses on building efficient backbones from the outset. This line of work is particularly motivated by the need to alleviate the high computational cost and memory bottlenecks introduced by self-attention. Representative examples include MobileViT~\cite{mehta2022mobilevit} and EfficientViT~\cite{liu2023efficientvit}, both of which actively exploit the fast computation and memory efficiency of CNNs, while restricting costly attention operations to only the necessary scope or reformulating them in a more efficient manner. MobileViT achieves high efficiency in mobile settings by adopting a hybrid CNN–Transformer architecture, where convolution is used to efficiently extract local features and the Transformer is employed to complement them with global contextual information. In contrast, EfficientViT retains a Transformer-based architecture while redesigning its building blocks to reduce redundant computation and memory inefficiency within the attention operation itself, thereby achieving both faster inference and higher accuracy. Overall, architectural lightweighting methods offer the advantage of improving efficiency at the backbone level; however, because they typically require designing and training new architectures from scratch, they are limited in their ability to be readily applied to the rapidly growing number of newly emerging pretrained models.

\subsection{Token Reduction Methods}
Research that directly reduces the number of tokens to lower computational cost exploits the redundancy inherent in the patch-sequence representation of ViTs~\cite{marin2021token, kim2022learned, yin2022vit}. An early representative work, DynamicViT~\cite{rao2021dynamicvit}, proposed a dynamic token pruning strategy in which a lightweight prediction module estimates token importance at each layer and progressively removes unimportant tokens. In contrast, EViT~\cite{liang2022evit} uses class-token attention to distinguish informative tokens from inattentive ones, and then reduces the computation of subsequent MHSA and FFN layers by reorganizing and fusing the latter into the important tokens. ToMe~\cite{bolya2023tome} further extended this line of work to training-free token merging, presenting a plug-and-play acceleration method for pretrained ViTs that progressively merges similar tokens based on token similarity.

Subsequent studies have explored combining the advantages of pruning and merging~\cite{kim2024token, wei2023joint}. ToFu, for example, selectively combines pruning and merging according to the functional linearity of each layer, thereby exploring a compromise between the two families of approaches. More recently, this line of research has been extended to vision-language models (VLMs), where plug-and-play token reduction methods have been proposed to improve the efficiency of visual token processing including SAINT and VisPruner~\cite{jeddi2025similarity, zhang2025beyond}.

\section{Methods}

\begin{figure*}[t]
    \centering
    \includegraphics[width=\textwidth]{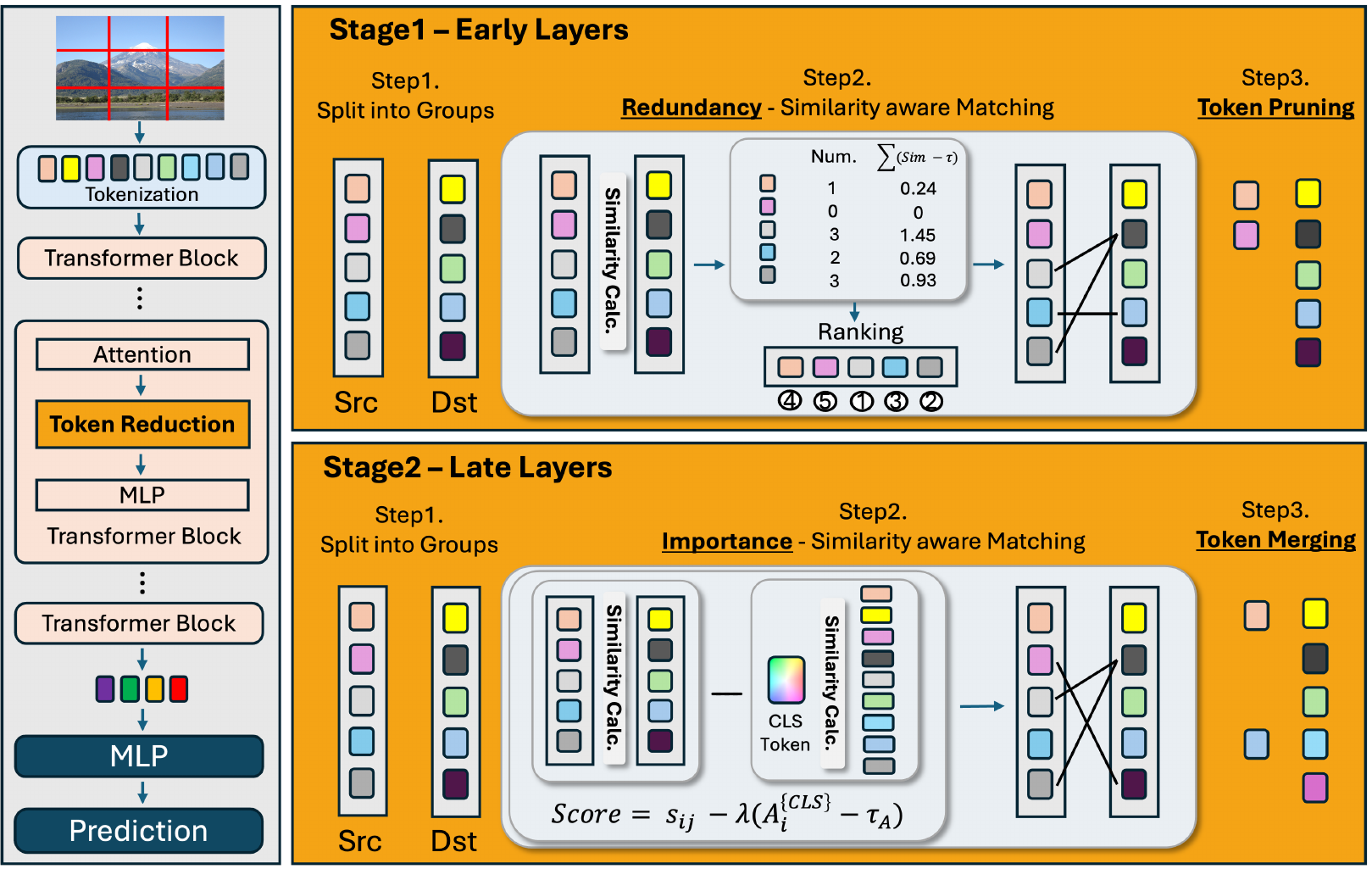}
    \caption{Overall architecture of the proposed algorithm. At each stage, the tokens are first divided into two groups. Then, in the early layers, token reduction is performed by ranking tokens using a redundancy-similarity aware metric, while in the late layers, token reduction is performed using an importance-similarity aware metric. Num. indicates the number of destination tokens whose similarity exceeds the threshold, and the equations shown represent the metrics used for ranking.}
    \label{fig:main}
\end{figure*}

Because the properties of Vision Transformer tokens vary substantially across layer depth, we argue that token compression algorithms should be designed in a way that explicitly accounts for this variation. Motivated by this perspective, we propose RAPID, a plug-and-play token reduction module that is tailored to the layer-wise characteristics of ViT tokens. By adapting to the representational properties of tokens at different depths, RAPID is designed to more effectively leverage the complementary strengths of token pruning and token merging.

\subsection{Redundancy-Similarity Aware Pruning}

As mentioned above, the shallow-to-middle layers of a Vision Transformer mainly learn local patterns. In addition, when token reduction is applied layer by layer, these layers contain substantially more tokens than the later layers. Therefore, a considerable number of tokens contain overlapping or highly similar information. Reflecting this property, we design the token reduction module at this stage as a token pruning algorithm that ranks patch tokens using a metric that jointly considers redundancy and similarity.

More specifically, the method first takes as input the number of tokens to be reduced at each layer, denoted by r. Given this setting, the patch tokens are divided into two groups according to their indices, namely the odd-indexed group and the even-indexed group, and the pairwise similarity between the two groups is computed~\cite{bolya2023tome}. Then, the predefined threshold value is subtracted from each pairwise similarity score. Based on the resulting values, positive values are retained, while values less than or equal to zero are set to zero. These values are then summed for each source token. This process is formulated as follows.

\begin{equation}
\mathrm{score}_i = \sum_j \max(s_{ij} - \tau, 0)
\end{equation}

Here, $s_{ij}$ denotes the pairwise similarity between source token $i$ and destination token $j$, and $\tau$ denotes the threshold.

This metric is designed to be mainly influenced by two factors. First, the final summation value increases as the number of positive values increases, that is, as more pairwise similarity values exceed the threshold. Second, the score increases as the similarity values themselves become larger. In particular, when the number of values exceeding the threshold is similar across tokens, the metric is designed so that larger individual similarity values have a greater influence on the final score.

By computing the score in this way, the model jointly reflects redundancy and similarity. As a result, it achieves faster processing while more effectively identifying tokens whose removal is least likely to harm performance, thereby minimizing the performance degradation caused by token reduction. Finally, based on the computed scores, the top $r$ source tokens with the highest values are pruned. The overall algorithm proceeds as follows.
\begin{enumerate}[label=\arabic*., leftmargin=*, itemsep=2pt, topsep=5pt, parsep=0pt, partopsep=0pt]
    \item Partition the remaining tokens into a source set $\mathbb{A}$ and a destination set $\mathbb{B}$ according to their token indices.

    \item Compute the pairwise cosine similarity between each source token in $\mathbb{A}$ and each destination token in $\mathbb{B}$.

    \item For each source token, subtract the similarity threshold from each pairwise similarity score and clamp negative values to zero.

    \item Sum these threshold-exceeding similarity margins over all destination tokens to obtain the redundancy score for each source token.

    \item Prune the top-$r$ source tokens with the highest redundancy scores.
\end{enumerate}

\subsection{Importance-Similarity Aware Merging}

Next, in the middle-to-late layers following the shallow-to-middle stages, tokens begin to capture more global context, and the CLS token which has a strong influence on the final prediction, also begins to form a genuinely meaningful representation that reflects semantic information. Accordingly, we design the algorithm in these middle-to-late layers to preserve semantically important tokens as much as possible. This aims to minimize the impact of token reduction on prediction while reducing the remaining, relatively less important tokens in a way that maintains diversity.

At this stage, we use the attention score of the CLS token to determine which tokens are important~\cite{liang2022evit}. First, we select the value corresponding to the top $n\%$ of the CLS-token attention scores and use this value as a threshold. Next, as in the Redundancy-Similarity Aware Pruning stage, the tokens are divided into a source token set and a destination token set, and a similarity matrix is computed between them. Then, the threshold value is subtracted from the CLS-attention score of the patch corresponding to each source token, and the resulting value is incorporated into each pairwise similarity score for that source token to obtain the final score. Since the CLS-attention score is generally much smaller than each pairwise similarity value, we introduce a hyperparameter $\lambda$ to adjust its influence on the final score computation and balance the contribution of the importance term. The score for each token pair is formulated as follows.

\begin{equation}
\mathrm{score}_{ij} = s_{ij} - \lambda \left(A_i^{\mathrm{CLS}} - \tau_A\right)
\label{eq:importance_score}
\end{equation}

\begin{equation}
\mathcal{C} = \left\{(i,j) \mid j=\arg\max_{j'} \mathrm{score}_{ij'} \right\}
\label{eq:candidate_pairs}
\end{equation}

\begin{equation}
\mathcal{M} = \operatorname{TopK}_{r}\!\left(\mathcal{C} \,;\, \mathrm{score}_{ij}\right)
\label{eq:Topk_merge}
\end{equation}

In Equation~\eqref{eq:importance_score}, $s_{ij}$ denotes the pairwise similarity between source token $i$ and destination token $j$. $A_i^{\mathrm{CLS}}$ denotes the CLS-attention score of source token $i$, $\tau_A$ denotes the threshold corresponding to the top $n\%$ of the CLS-attention scores, and $\mathrm{score}_{ij}$ denotes the importance-similarity aware score between source token $i$ and destination token $j$. Based on these scores, $\mathcal{C}$ is constructed as the candidate merge-pair set by matching each source token to its highest-scoring destination token, as shown in Equation~\eqref{eq:candidate_pairs}. The final merge set $\mathcal{M}$ is then obtained by selecting the top-$r$ candidate pairs from $\mathcal{C}$ according to $\mathrm{score}_{ij}$, as defined in Equation~\eqref{eq:Topk_merge}.

Unlike in the shallow-to-middle layers, where token pruning is used, in the middle-to-late layers we adopt token merging to fuse the selected token pairs~\cite{kim2024token}. Specifically, for each source token, we select the destination token with the highest $\mathrm{score}_{ij}$ value and use this maximum score to rank the source tokens. The top $r$ source tokens with the highest maximum scores are selected for merging with their corresponding destination tokens, while the remaining source tokens are left unmerged. The reason is that, at this stage of the network, each token has already acquired a distinctive semantic meaning through repeated transformations across layers. Therefore, pruning would completely discard the information carried by a token, whereas merging highly similar tokens makes it possible to preserve, at least to some extent, the semantic content of both. In addition, from a representation perspective, when two similar tokens are fused in the middle-to-late layers, the resulting token is more likely to behave as an intermediate point between the two original representations. The overall algorithm proceeds as follows.
\begin{enumerate}[label=\arabic*., leftmargin=*, itemsep=2pt, topsep=5pt, parsep=0pt, partopsep=0pt]
    \item Compute CLS-attention scores for patch tokens and determine the cutoff value $\tau_A$ corresponding to the top-$n\%$ of candidate image tokens.

    \item Partition the remaining tokens into a source set $\mathbb{A}$ and Compute the pairwise cosine similarity between each source token in $\mathbb{A}$ and each destination token in $\mathbb{B}$.

    \item For each source token, compute an importance offset by subtracting the cutoff value $\tau_A$ from its CLS-attention score. 

    \item The pairwise merge score is then obtained by subtracting this offset from the cosine similarity score, with a weighting factor $\lambda$ controlling the strength of the importance penalty.

    \item Based on the adjusted merge scores, identify the most suitable destination token for each source token. Then, rank the source tokens according to their highest adjusted merge scores and select the top-$r$ source tokens for merging.

    \item Merge the selected source tokens into their matched destination tokens.
\end{enumerate}

\section{Experiments and Results}

\begin{table*}[t]
    \caption{Comparison of Accuracy and Throughput on ImageNet-1K across various backbones and reduction rates ($r$) using plug-and-play token reduction modules. For ViT-L, token reduction is applied only to the first 12 layers using the same layer-wise schedule as ViT-B and DeiT-B.}
    \label{tab:vit_drop_comparison}
    \centering
    \small
    \setlength{\tabcolsep}{5pt}
    \renewcommand{\arraystretch}{1.05}

    \begin{tabular}{
        >{\centering\arraybackslash}m{1.3cm}
        >{\centering\arraybackslash}m{1.5cm}
        *{8}{>{\centering\arraybackslash}m{1.05cm}}
    }
        \toprule
        \multirow{3}{*}{\textbf{Backbone}} & \multirow{3}{*}{\textbf{Method}}
        & \multicolumn{2}{c}{$r=11$}
        & \multicolumn{2}{c}{$r=14$}
        & \multicolumn{2}{c}{$r=17$}
        & \multicolumn{2}{c}{$r=20$} \\

        & 
        & \multicolumn{2}{c}{\scriptsize (output tokens = 65)}
        & \multicolumn{2}{c}{\scriptsize (output tokens = 29)}
        & \multicolumn{2}{c}{\scriptsize (output tokens = 8)}
        & \multicolumn{2}{c}{\scriptsize (output tokens = 4)} \\
        
        \cmidrule(lr){3-4}\cmidrule(lr){5-6}\cmidrule(lr){7-8}\cmidrule(lr){9-10}
        & & \textbf{Accuracy} & \textbf{Thru.}
          & \textbf{Accuracy} & \textbf{Thru.}
          & \textbf{Accuracy} & \textbf{Thru.}
          & \textbf{Accuracy} & \textbf{Thru.} \\
        \midrule

        \multirow{4}{*}{DeiT-B}
        & Full Model & 81.97 & 1938 & 81.97 & 1938 & 81.97 & 1938 & 81.97 & 1938 \\
        \noalign{\vskip 2pt}
        \cdashline{2-10}
        \noalign{\vskip 2pt}
        & ToMe~\cite{bolya2023tome}       & 80.59 & 2693 & 79.42 & 3104 & 76.38 & 3553 & 66.47 & 3998 \\
        & ToFu~\cite{kim2024token}       & 80.98 & 2710 & 80.23 & 3128 & 78.24 & 3586 & 71.03 & 4038 \\        
        & \textbf{Ours} & \textbf{81.25} & 2653 & \textbf{80.52} & 3052 & \textbf{78.77} & 3492 & \textbf{72.03} & 3926 \\
        \midrule

        \multirow{4}{*}{ViT-B}
        & Full Model & 85.11 & 1661 & 85.11 & 1661 & 85.11 & 1661 & 85.11 & 1661 \\
        \noalign{\vskip 2pt}
        \cdashline{2-10}
        \noalign{\vskip 2pt}
        & ToMe       & 84.13 & 2688 & 83.41 & 3098 & 81.30 & 3544 & 73.25 & 3987 \\
        & ToFu       & 84.25 & 2714 & 83.72 & 3127 & 82.02 & 3587 & 75.26 & 4038 \\
        & \textbf{Ours} & \textbf{84.61} & 2656 & \textbf{84.17} & 3053 & \textbf{82.83} & 3495 & \textbf{77.54} & 3930 \\
        \midrule

        \multirow{4}{*}{ViT-L}
        & Full Model & 85.85 & 621 & 85.85 & 621 & 85.85 & 621 & 85.85 & 621 \\
        \noalign{\vskip 2pt}
        \cdashline{2-10}
        \noalign{\vskip 2pt}
        & ToMe       & 80.88 & 1170 & 61.36 & 1567 & 3.96 & 1946 & 0.35 & 2193 \\
        & ToFu       & 82.58 & 1175 & 68.63 & 1585 & 6.59 & 1955 & 0.50 & 2222 \\
        & \textbf{Ours} & \textbf{82.98} & 1159 & \textbf{69.64} & 1541 & \textbf{7.15} & 1924 & \textbf{0.52} & 2183 \\
        \bottomrule
    \end{tabular}
\end{table*}

\subsection{Experimental Setup}

In this experiment, we evaluate the performance of the proposed method using ImageNet-1K pretrained Vision Transformer backbones provided by the timm library~\cite{rw2019timm}. The experiments are conducted on models of various scales, including ViT-Base, ViT-Large, and DeiT-Base. All evaluations are performed on the ImageNet-1K validation set with a batch size of 64~\cite{deng2009imagenet}. After applying the proposed method to each backbone, we analyze the performance variation by changing the reduction parameter \(r\)~\cite{bolya2023tome, kim2024token}.

For the hyperparameter settings, we set the similarity threshold used in redundancy-similarity aware pruning to 0.7. In importance-similarity aware merging, the importance threshold and the weighting factor $\lambda$ are set to 20\% and 50, respectively. These hyperparameter values are fixed across all backbone models and all reduction settings to ensure a consistent evaluation protocol.

Accuracy and memory usage are measured during inference over the entire validation set. During inference, we use automatic mixed precision(AMP) with FP16 autocasting to improve computational efficiency. Here, peak memory refers to the maximum amount of GPU memory actually allocated during the entire inference process. Meanwhile, to measure throughput while excluding the effects of data loading and accuracy computation, we use synthetic inputs of size \(3 \times 224 \times 224\). After a sufficient warm-up period, multiple forward passes are repeatedly performed, and throughput is calculated based on the average execution time. The results are reported in terms of the number of images processed per second. All experiments are conducted on a single NVIDIA RTX 4090 GPU.

\subsection{Main Results}
Table~1 shows that the proposed method consistently achieves higher accuracy than ToMe and ToFu across different backbones, especially as the reduction parameter $r$ increases. Although our method yields slightly lower throughput under the same reduction
setting, it preserves higher Top-1 accuracy across all cases, leading to a more favorable accuracy-efficiency trade-off in high-compression regimes. This suggests that the proposed strategy better preserves informative token representations under the same token budget, thereby effectively mitigating performance degradation, especially
under aggressive token reduction.
For ViT-L, which is twice as deep as the base-scale backbones, using the same layer-wise reduction schedule concentrates token reduction in the earlier portion of the network, causing accuracy to drop sharply under large $r$ settings. Nevertheless, even in this extreme setting, our method still achieves higher accuracy than ToMe and ToFu, demonstrating its relative robustness to aggressive token reduction.

\begin{figure}[t]
    \centering
    \includegraphics[width=\linewidth]{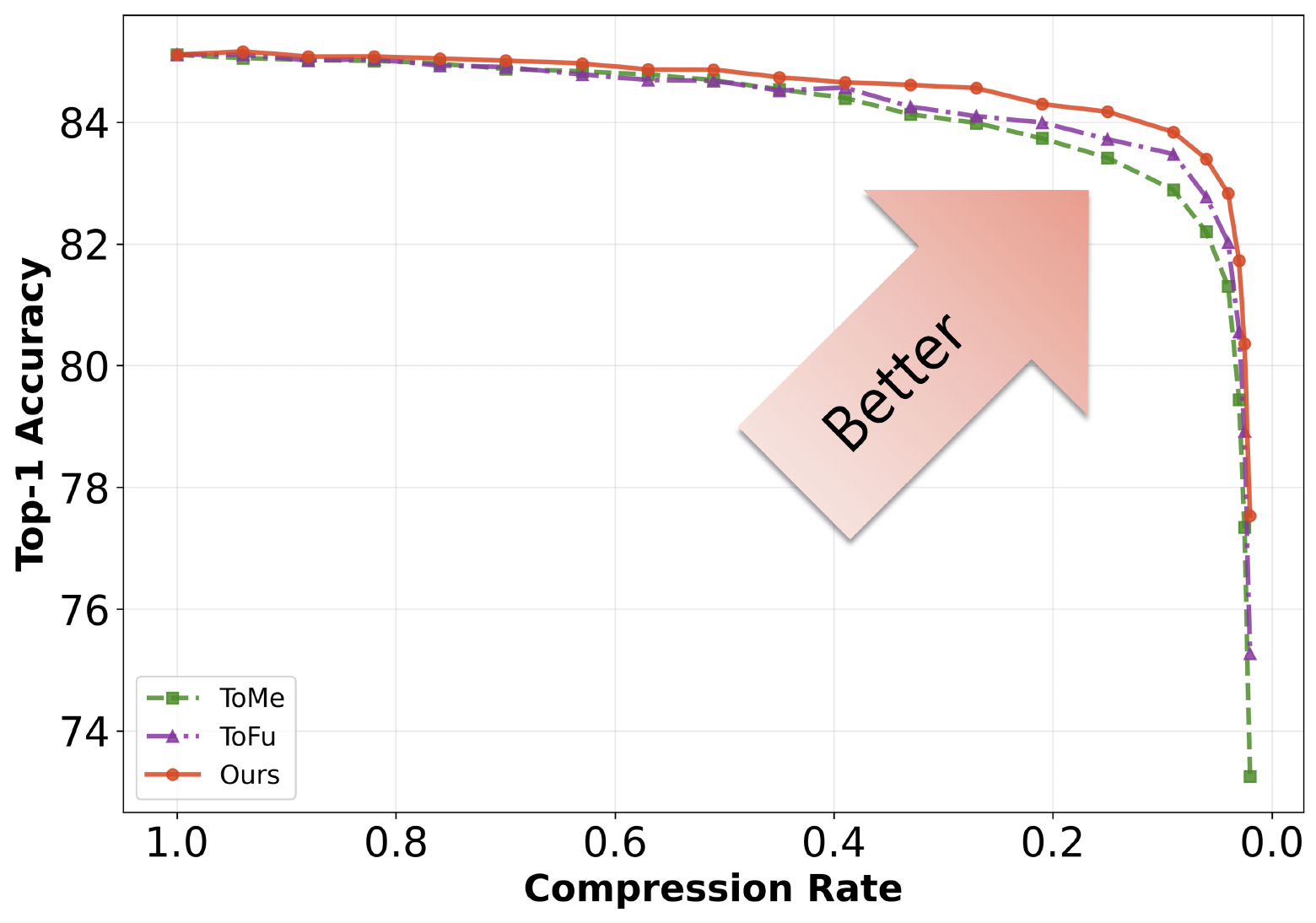}
    
    \vspace{0.0em}
    
    \includegraphics[width=\linewidth]{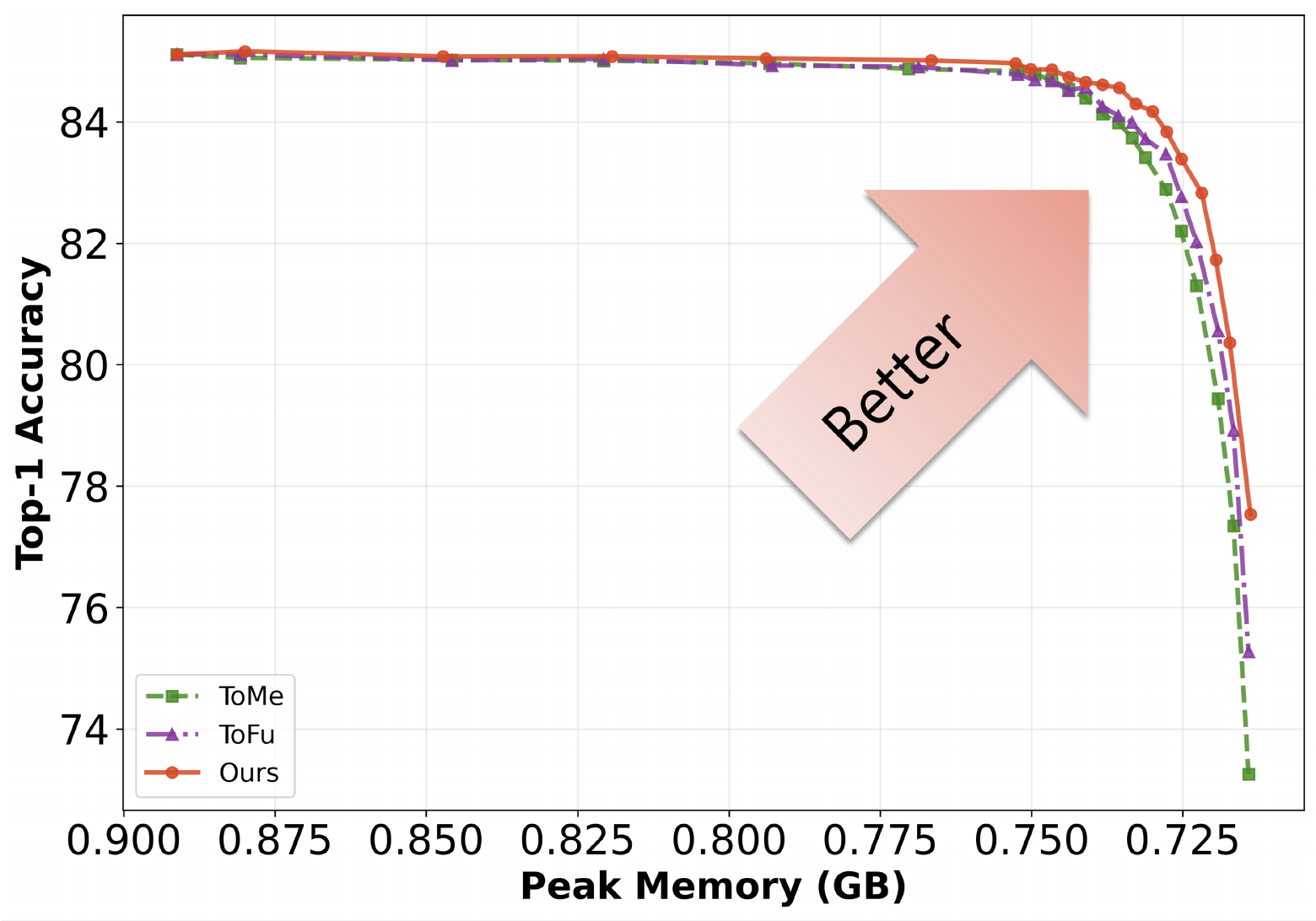}
    
    \caption{Efficiency-Accuracy trade-off comparison of ToMe, ToFu, and Ours on ViT-B. The top plot shows compression rate versus Top-1 accuracy, and the bottom plot shows peak memory (GB) versus Top-1 accuracy. A lower compression rate indicates stronger token reduction.}
    \label{fig:throughput_tradeoff}
\end{figure}

Figure~2 further illustrates the accuracy-efficiency trade-off under the ViT-B setting. When evaluated at comparable compression rates and peak memory usage, the proposed method generally maintains higher Top-1 accuracy than representative plug-and-play token reduction methods, including ToMe and ToFu. This suggests that the proposed method can preserve useful visual information more effectively under a similar computational and memory budget. While the improvement is not intended to indicate a large efficiency gain in every operating region, the results show that the proposed strategy provides a consistently favorable performance by achieving slightly better accuracy at comparable levels of token compression and memory usage.

\begin{table}[t]
\centering
\caption{Comparison of Top-1 accuracy and compression rate across methods using DeiT-S as the backbone. Acc. denotes Top-1 accuracy, and Comp. Rate denotes the percentage of retained output tokens relative to the original input tokens. Methods below the dashed line are directly applicable without additional training.}
\label{tab:comparison}
\begin{tabular*}{\linewidth}{@{\extracolsep{\fill}}clcc}
\hline
Plug-and-Play & Method & Acc. & Comp. Rate (\%) \\
\hline
X & DynamicViT~\cite{rao2021dynamicvit} & 79.32 & 35 \\
X & SP-ViT~\cite{kong2022spvit}     & 79.34 & 35 \\
X & EViT~\cite{liang2022evit}& 79.36 & 35 \\
\cdashline{1-4}
O & ToMe~\cite{bolya2023tome}       & 79.25 & 35 \\
O & ToFu~\cite{kim2024token}       & 79.08 & 35 \\
O & Ours                           & \textbf{79.57} & 35 \\
\hline
\end{tabular*}
\end{table}

Table~2 compares the proposed method with representative token reduction methods on the DeiT-S~\cite{touvron2021training} backbone. For methods that require fine-tuning, including DynamicViT, SP-ViT, and EViT, we used the corresponding pretrained weights provided by the original methods. In contrast, ToMe, ToFu, and the proposed method are plug-and-play approaches that can be directly applied to a pretrained backbone without additional fine-tuning, making them more flexible in practice. Although these plug-and-play methods do not rely on task-specific fine-tuning, the proposed method achieves competitive accuracy relative to fine-tuning-based approaches under the same 35\% compression rate. In particular, our method records the highest Top-1 accuracy among the compared methods, achieving 79.57\% accuracy at a compression rate of 35\%. For ToFu, since the official implementation is not publicly available, we implemented the method based on the algorithm described in the paper and evaluated it under the same setting. These results indicate that the proposed method provides an effective balance between applicability and performance, showing strong accuracy relative to the retained token ratio while preserving the flexibility of a plug-and-play token reduction module.

\subsection{Ablation Study}

\begin{table}[t]
\centering
\caption{Performance comparison under different merge configurations on \texttt{ViT-B} with $r=20$. Red. denotes redundancy aware pruning strategy, and Imp. denotes importance aware merging strategy.}
\label{tab:merge_config_r22}
\setlength{\tabcolsep}{3.5pt}
\renewcommand{\arraystretch}{1.05}
\begin{tabularx}{\columnwidth}{c c X c c}
\toprule
Top-1 & Top-5 & \centering Configuration & Red. & Imp. \\
\midrule
76.26 & 92.73 & \centering R I I I I I I I I I I I  & 1  & 11 \\
76.94 & 93.11 & \centering R R I I I I I I I I I I & 2  & 10 \\
77.28 & 93.29 & \centering R R R I I I I I I I I I & 3  & 9  \\
\textbf{77.54} & \textbf{93.51} & \centering R R R R I I I I I I I I & 4  & 8  \\
77.22 & 93.36 & \centering R R R R R I I I I I I I & 5  & 7  \\
76.97 & 93.12 & \centering R R R R R R I I I I I I & 6  & 6  \\
75.66 & 92.60 & \centering R R R R R R R I I I I I & 7  & 5  \\
73.77 & 91.49 & \centering R R R R R R R R I I I I & 8  & 4  \\
70.01 & 88.87 & \centering R R R R R R R R R I I I & 9  & 3  \\
66.17 & 86.11 & \centering R R R R R R R R R R I I & 10 & 2  \\
65.55 & 85.74 & \centering R R R R R R R R R R R I & 11 & 1  \\
\bottomrule
\end{tabularx}
\end{table}

A key aspect of the proposed method is that it does not apply a single token reduction strategy uniformly across all layers; instead, it employs different reduction algorithms according to the representational characteristics of each layer. To validate this design choice, we conducted an ablation study to analyze at which layer switching the reduction strategy is most effective. In Table 3, R denotes a pruning-based reduction method that reduces tokens by jointly considering redundancy and token similarity, whereas I denotes a merging-based method that reflects both token importance and similarity.

According to the results in Table 3, the configuration that switches the reduction strategy after the 4th layer achieves the best performance. This setting records 77.54 Top-1 accuracy and 93.51 Top-5 accuracy, yielding the best result among all tested configurations. This suggests that redundancy-centered pruning is more effective in the shallow-to-middle stages of the network, whereas importance-centered merging becomes more appropriate beyond a certain depth. 

In addition, as the switching point is moved later than the 4th layer, performance shows an overall tendency to decrease in a relatively linear yet substantial manner. This implies that, in the later layers, as the network depth increases, token representations tend to encode higher-level semantic information and discriminative importance rather than simple local redundancy. Therefore, integrating tokens while preserving important information is far more appropriate than merely removing redundant tokens.

Otherwise, applying importance-aware merging too early also fails to achieve optimal performance. This indicates that, in the early layers, token representations are still dominated by low-level visual patterns and local structures; at this stage, redundancy-based elimination may be more effective than importance-based integration. Since the early and intermediate layers contain a relatively large amount of similar or redundant tokens, reducing this redundancy first can be more beneficial for subsequent information processing. Therefore, if the strategy shifts to merging too early, representations that have not yet been sufficiently refined may be fused prematurely, making it more difficult to preserve information optimally.

\section{Conclusion}

In this paper, we proposed RAPID, a depth-aware token reduction framework that applies different reduction strategies by reflecting the layer-wise characteristics of token representations in Vision Transformers. Our method is motivated by the observation that the nature of tokens within a ViT changes substantially with depth. Accordingly, in the shallow-to-middle layers, we apply redundancy-aware pruning to reflect the prevalence of local and redundant visual patterns, whereas in the middle-to-late layers, we apply importance-aware merging in order to preserve tokens that carry more semantic and prediction-critical information. By jointly considering not only similarity, but also redundancy and importance during the token reduction process, the proposed framework is designed to enable more effective token compression while minimizing information loss.

The experimental results presented above clearly demonstrate the effectiveness of RAPID on image classification tasks across a variety of backbone models. Although further investigation is needed to determine whether the proposed framework can also be effectively applied to more computationally demanding and accuracy-sensitive tasks such as segmentation and generation, we believe that this algorithm has strong potential to be used effectively in a wide range of applications that require lightweight Vision Transformers. Another promising direction for future work is to establish a more principled theoretical criterion for determining the layer depth at which the reduction strategy should transition from pruning to merging. Such a criterion could provide a deeper understanding of the layer-wise evolution of token representations and further enhance the robustness and applicability of depth-aware token reduction frameworks.

\balance
\bibliographystyle{ACM-Reference-Format}
\bibliography{Ref}

\end{document}